\documentclass{article} 
\usepackage{nips15submit_e,times}
\usepackage{hyperref}
\usepackage{url}

\usepackage{graphicx}
\usepackage{amsmath}

\graphicspath{{figures/}}

\title{Training Spiking Deep Networks\\for Neuromorphic Hardware}

\author{
Eric Hunsberger \\
Centre for Theoretical Neuroscience \\
University of Waterloo \\
Waterloo, ON N2L 3G1 \\
\texttt{ehunsber@uwaterloo.ca} \\
\And
Chris Eliasmith \\
Centre for Theoretical Neuroscience \\
University of Waterloo \\
Waterloo, ON N2L 3G1 \\
\texttt{celiasmith@uwaterloo.ca} \\
}

\nipsfinalcopy 

\begin{document}

\maketitle

\begin{abstract}
We describe a method to train spiking deep networks
that can be run using leaky integrate-and-fire (LIF) neurons,
achieving state-of-the-art results for spiking LIF networks on five datasets,
including the large ImageNet ILSVRC-2012 benchmark.
Our method for transforming deep artificial neural networks
into spiking networks is scalable
and works with a wide range of neural nonlinearities.
We achieve these results by softening the neural response function,
such that its derivative remains bounded,
and by training the network with noise
to provide robustness against the variability introduced by spikes.
Our analysis shows that implementations of these networks
on neuromorphic hardware will be many times more power-efficient
than the equivalent non-spiking networks on traditional hardware.
\end{abstract}

\section{Introduction}

Deep artificial neural networks (ANNs) have recently been very successful
at solving image categorization problems.
Early successes with the MNIST database \cite{Lecun1998}
were subsequently tested on the more difficult but similarly sized
CIFAR-10~\cite{Krizhevsky2010} and
Street-view house numbers~\cite{Sermanet2012} datasets.
Recently, many groups have achieved better results
on these small datasets (e.g. \cite{Lee2015}),
as well as on larger datasets (e.g. \cite{Gens2012}).
This work has culminated with the application of
deep convolutional neural networks to ImageNet \cite{Krizhevsky2012},
a very large and challenging dataset
with 1.2 million images across 1000 categories.

There has recently been considerable effort
to introduce neural ``spiking'' into deep
ANNs~\cite{Eliasmith2012a, Neftci2013, O'Connor2013, Diehl2015, Cao2014, Esser2016},
such that connected nodes in the network transmit information
via instantaneous single bits (spikes),
rather than transmitting real-valued activities.
While one goal of this work is to better understand the brain
by trying to reverse engineer it \cite{Eliasmith2012a},
another goal is to build energy-efficient neuromorphic systems
that use a similar spiking communication method,
for image categorization \cite{Diehl2015, Cao2014, Esser2016}
or other applications \cite{Diehl2016}.

In this paper, we present a novel method for translating deep ANNs
into spiking networks for implementation on neuromorphic hardware.
Unlike previous methods,
our method is applicable to a broad range of neural nonlinearities,
allowing for implementation on hardware with idiosyncratic neuron types (e.g.~\cite{Benjamin2014}).
We extend our previous results~\cite{Hunsberger2015} to additional datasets,
and most notably demonstrate that it scales to the large ImageNet dataset.
We also perform an analysis demonstrating that
neuromorphic implementations of these networks will be
many times more power-efficient than
the equivalent non-spiking networks running on traditional hardware.

\section{Methods}

We first train a network on static images
using traditional deep learning techniques; we call this the ANN.
We then take the parameters (weights and biases) from the ANN
and use them to connect spiking neurons,
forming the spiking neural network (SNN).
A central challenge is to train the ANN in such a way that
it \emph{can} be transferred into a spiking network,
and such that the classification error of the resulting SNN is minimized.

\subsection{Convolutional ANN}

We base our network off that of Krizhevsky et al. \cite{Krizhevsky2012},
which won the ImageNet ILSVRC-2012 competition.
A smaller variant of the network achieved ~11\% error on the CIFAR-10 dataset.
The network makes use of a series of \emph{generalized convolutional layers},
where one such layer is composed of a set of convolutional weights,
followed by a neural nonlinearity, a pooling layer,
and finally a local contrast normalization layer.
These generalized convolutional layers are followed by either locally-connected layers,
fully-connected layers, or both, all with a neural nonlinearity.
In the case of the original network,
the nonlinearity is a rectified linear (ReLU) function,
and pooling layers perform max-pooling.
The details of the network can be found in~\cite{Krizhevsky2012}
and code is available\footnote{\url{https://github.com/akrizhevsky/cuda-convnet2}}.

To make the ANN transferable to spiking neurons,
a number of modifications are necessary.
First, we remove the local response normalization layers.
This computation would likely require some sort of lateral connections
between neurons,
which are difficult to add in the current framework
since the resulting network would not be feedforward
and we are using methods focused on training feedforward networks.

Second, we changed the pooling layers from max pooling to average pooling.
Again, computing max pooling would likely require lateral connections
between neurons,
making it difficult to implement without significant changes
to the training methodology.
Average pooling, on the other hand, is very easy to compute in spiking neurons,
since it is simply a weighted sum.

The other modifications---%
using leaky integrate-and-fire neurons and training with noise---%
are the main focus of this paper,
and are described in detail below.

\subsection{Leaky integrate-and-fire neurons}

Our network uses a modified leaky integrate-and-fire (LIF) neuron nonlinearity
instead of the rectified linear nonlinearity.
Past work has kept the rectified linear nonlinearity for the ANN
and substituted in the spiking integrate-and-fire (IF) neuron model
in the SNN \cite{Cao2014, Diehl2015},
since the static firing curve of the IF neuron model is a rectified line.
Our motivation for using the LIF neuron model is that it
and it demonstrates that more complex, nonlinear neuron models
can be used in such networks.
Thus, these methods can be extended to the idiosyncratic neuron types
employed by some neuromorphic hardware (e.g. \cite{Benjamin2014}).

The LIF neuron dynamics are given by the equation
\begin{align}
  \tau_{RC} \dot v(t) = -v(t) + J(t)
  \label{eqn:lifode}
\end{align}
where $v(t)$ is the membrane voltage,
$\dot v(t)$ is its derivative with respect to time,
$J(t)$ is the input current,
and $\tau_{RC}$ is the membrane time constant.
When the voltage reaches $V_{th} = 1$, the neuron fires a spike,
and the voltage is held at zero for a refractory period of $\tau_{ref}$.
Once the refractory period is finished,
the neuron obeys Equation~\ref{eqn:lifode} until another spike occurs.

Given a constant input current $J(t) = j$,
we can solve Equation~\ref{eqn:lifode} for the time it takes the voltage
to rise from zero to one,
and thereby find the steady-state firing rate
\begin{align}
  r(j) = \left[
    \tau_{ref} + \tau_{RC} \log\left(
        1 + \frac{V_{th}}{\rho(j - V_{th})}\right)
    \right]^{-1}
  \label{eqn:lifss}
\end{align}
where $\rho(x) = \max(x, 0)$.

Theoretically, we should be able to train a deep neural network
using Equation~\ref{eqn:lifss} as the static nonlinearity
and make a reasonable approximation of the network in spiking neurons,
assuming that the spiking network has a synaptic filter that sufficiently
smooths a spike train to give a good approximation of the firing rate.
The LIF steady state firing rate has the particular problem
that the derivative approaches infinity as $j \to 0_{+}$,
which causes problems when employing backpropagation.
To address this, we added smoothing to the LIF rate equation.

\begin{figure}
  \centering
  \includegraphics[width=0.7\columnwidth]{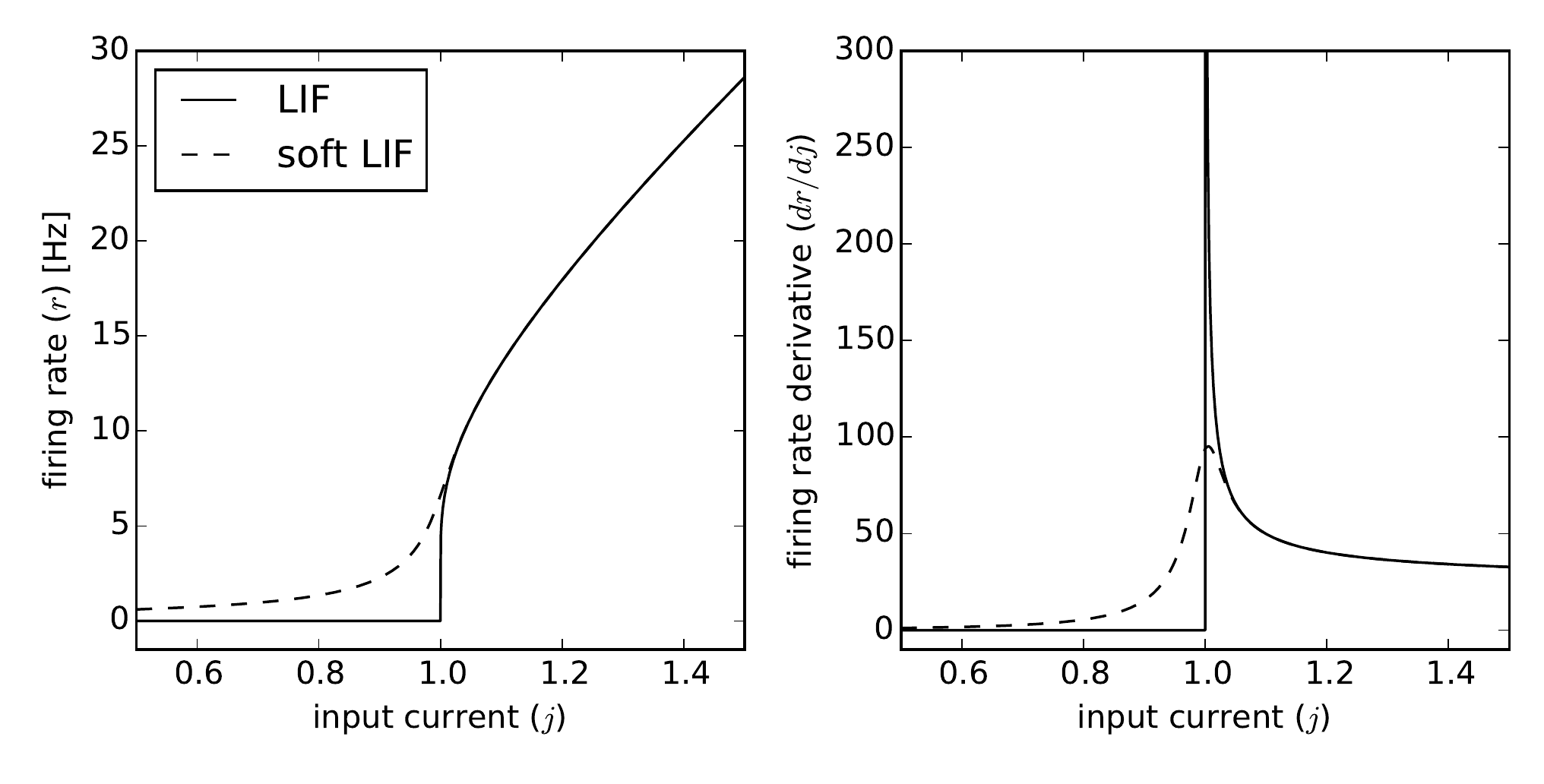}
  \caption{
    Comparison of LIF and soft LIF response functions.
    The left panel shows the response functions themselves.
    The LIF function has a hard threshold at $j = V_{th} = 1$;
    the soft LIF function smooths this threshold.
    The right panel shows the derivatives of the response functions.
    The hard LIF function has a discontinuous and unbounded derivative
    at $j = 1$; the soft LIF function has a continuous bounded derivative,
    making it amenable to use in backpropagation.
  }
  \label{fig:softlif}
\end{figure}

If we replace the hard maximum $\rho(x) = \max(x, 0)$
with a softer maximum $\rho_1(x) = \log(1 + e^x)$,
then the LIF neuron loses its hard threshold
and the derivative becomes bounded.
Further, we can use the substitution
\begin{align}
  \rho_2(x) = \gamma \log\left[1 + e^{x / \gamma}\right]
  \label{eqn:softrelusigma}
\end{align}
to allow us control over the amount of smoothing,
where $\rho_2(x) \to \max(x, 0)$ as $\gamma \to 0$.
Figure~\ref{fig:softlif} shows the result of this substitution.

\subsection{Training with noise}

Training neural networks with various types of noise on the inputs
is not a new idea.
Denoising autoencoders \cite{Vincent2008} have been successfully applied
to datasets like MNIST,
learning more robust solutions with lower generalization error than
their non-noisy counterparts.

In a biological spiking neural network,
synapses between neurons perform some measure of filtering on the spikes,
due to the fact that the post-synaptic current induced
by the neurotransmitter release is distributed over time.
We employ a similar mechanism in our networks to attenuate some of the
variability introduced by spikes.
The $\alpha$-function $\alpha(t) = (t / \tau_s) e^{-t / \tau_s}$
is a simple second-order lowpass filter,
inspired by biology~\cite{Mainen1995}.
We chose this as a synaptic filter for our networks since
it provides better noise reduction than a first-order lowpass filter.

The filtered spike train can be viewed as an estimate of the neuron activity.
For example, if the neuron is firing regularly at 200 Hz,
filtering spike train will result in a signal fluctuating around 200 Hz.
We can view the neuron output as being 200 Hz,
with some additional ``noise'' around this value.
By training our ANN with some random noise
added to the output of each neuron for each training example,
we can simulate the effects of using spikes
on the signal received by the post-synaptic neuron.

\begin{figure}
  \centering
  \includegraphics[width=0.55\columnwidth]{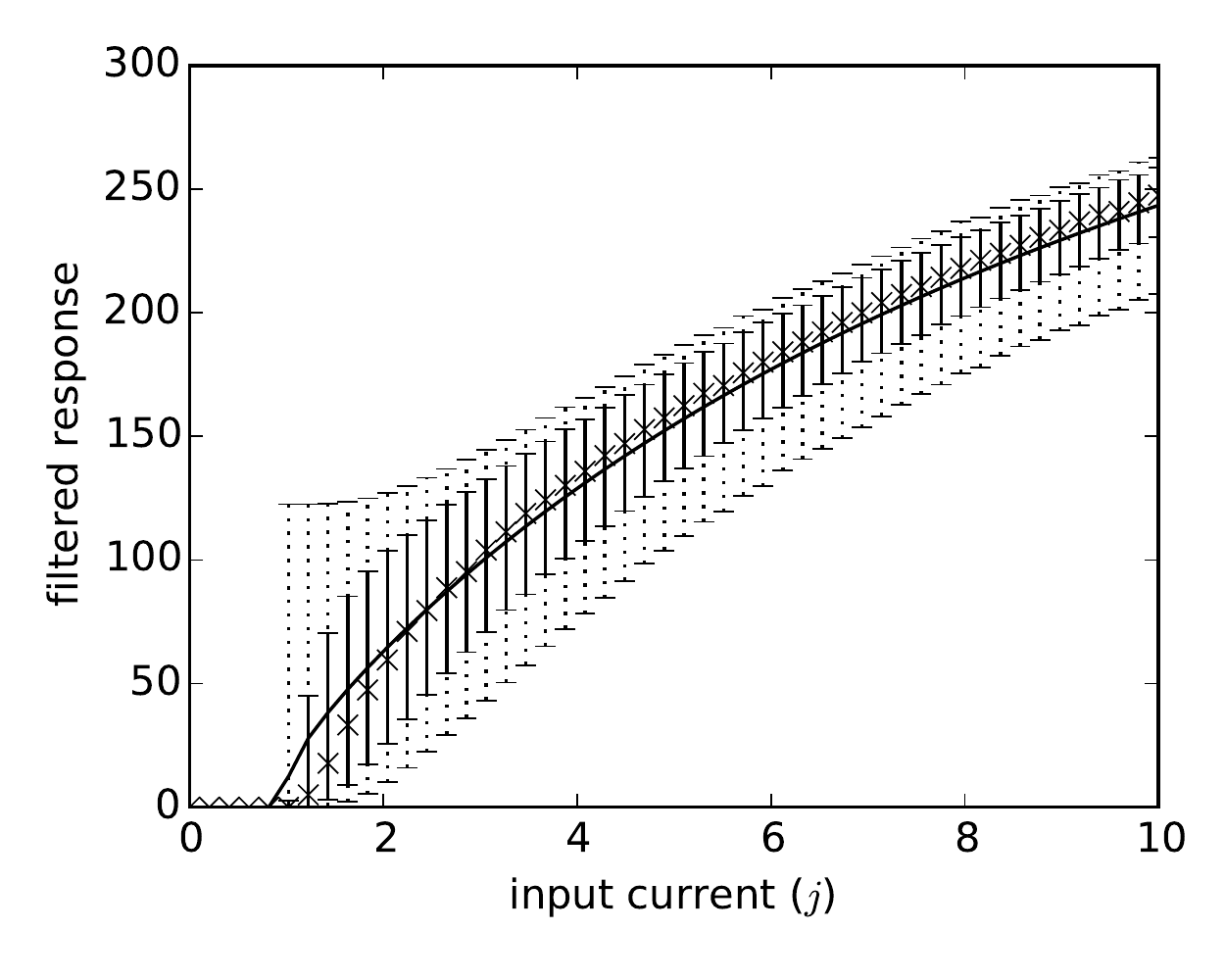}
  \caption{Variability in filtered spike trains versus
    input current for the LIF neuron
    ($\tau_{RC} = 0.02, \tau_{ref} = 0.004$).
    The solid line shows the mean of the filtered spike train
    (which matches the analytical rate of Equation~\ref{eqn:lifss}),
    the `x'-points show the median, the solid error bars show the 25th and 75th
    percentiles,
    and the dotted error bars show the minimum and maximum.
    The spike train was filtered with an $\alpha$-filter with $\tau_s = 0.003$ s.
  }
  \label{fig:noise}
\end{figure}

Figure~\ref{fig:noise} shows how the variability of filtered spike trains
depends on input current for the LIF neuron.
Since the impulse response of the $\alpha$-filter has an integral of one,
the mean of the filtered spike trains is equal to the analytical rate
of Equation~\ref{eqn:lifss}.
However, the statistics of the filtered signal vary significantly
across the range of input currents.
Just above the firing threshold,
the distribution is skewed towards higher firing rates
(i.e. the median is below the mean),
since spikes are infrequent so the filtered signal has time to return
to near zero between spikes.
At higher input currents, on the other hand,
the distribution is skewed towards lower firing rates
(i.e. the median is above the mean).
In spite of this,
we used a Gaussian distribution to generate the additive noise during training,
for simplicity.
We found the average standard deviation to be approximately $\sigma = 10$
across all positive input currents for an $\alpha$-filter with $\tau_s = 0.005$.
During training, we add Gaussian noise $\eta \sim G(0, \sigma)$
to the firing rate $r(j)$ (Equation~\ref{eqn:lifss})
when $j > 0$, and add no noise when $j \le 0$.

\subsection{Conversion to a spiking network}

Finally, we convert the trained ANN to a SNN.
The parameters in the spiking network (i.e. weights and biases)
are all identical to that of the ANN.
The convolution operation also remains the same,
since convolution can be rewritten as simple connection weights (synapses)
$w_{ij}$ between pre-synaptic neuron $i$ and post-synaptic neuron $j$.
(How the brain might \emph{learn} connection weight patterns, i.e. filters,
that are repeated at various points in space,
is a much more difficult problem that we will not address here.)
Similarly, the average pooling operation can be written
as a simple connection weight matrix,
and this matrix can be multiplied by the convolutional weight matrix
of the following layer to get direct connection weights between
neurons.\footnote{For computational efficiency,
we actually compute the convolution and pooling separately.}

The only component of the network that changes
when moving from the ANN to the SNN
is the neurons themselves.
The most significant change is that we replace
the soft LIF rate model (Equation~\ref{eqn:lifss})
with the LIF spiking model (Equation~\ref{eqn:lifode}).
We remove the additive Gaussian noise used in training.
We also add post-synaptic filters to the neurons,
which removes a significant portion of the high-frequency variation
produced by spikes.

\section{Results}

We tested our methods on five datasets:
MNIST~\cite{Lecun1998}, SVHN~\cite{Netzer2011},
CIFAR-10 and CIFAR-100~\cite{Krizhevsky2009},
and the large ImageNet ILSVRC-2012 dataset~\cite{Russakovsky2015}.
Our best result for each dataset is shown in Table~\ref{tab:results}.
Using our methods has allowed us to build spiking networks that
perform nearly as well as their non-spiking counterparts
using the same number of neurons.
All datasets show minimal loss in accuracy when transforming
from the ANN to the SNN.
\footnote{The ILSVRC-2012 dataset actually shows a marginal increase in accuracy,
though this is likely not statistically significant
and could be because the spiking LIF neurons have harder firing thresholds
than their soft-LIF rate counterparts.
Also, the CIFAR-100 dataset shows a considerable increase in performance
when using soft-LIF neurons versus ReLUs in the ANN,
but this could simply be due to the training hyperparameters chosen,
since these were not optimized in any way.}

\begin{table}
  \centering
  \begin{minipage}{\columnwidth}
    \begin{center}
    \begin{tabular}{|l|c|c|c|}\hline
      Dataset & ReLU ANN & LIF ANN & \textbf{LIF SNN} \\\hline\hline
      MNIST & 0.79\% & 0.84\% & 0.88\% \\\hline
      SVHN & 5.65\% & 5.79\% & 6.08\% \\\hline
      CIFAR-10 & 16.48\% & 16.28\% & 16.46\% \\\hline
      CIFAR-100 & 50.05\% & 44.35\% & 44.87\% \\\hline
      ILSVRC-2012 & 45.4\% (20.9\%)$^a$ & 48.3\% (24.1\%)$^a$ & 48.2\% (23.8\%)$^a$\\\hline
    \end{tabular}
    \end{center}
    \vspace{-0.2em}
    \hspace{5em}$^a$ Results from the first 3072-image test batch.
  \end{minipage}
  \vspace{0.2em}
  \caption{
    Results for spiking LIF networks (LIF SNN),
    compared with ReLU ANN and LIF ANN (both using the same network structure,
    but with ReLU and LIF rate neurons respectively).
    The spiking versions of each network perform
    almost as well as the rate-based versions.
    The ILSVRC-2012 (ImageNet) results show the error for the top result,
    with the top-5 result in brackets.
  }
  \label{tab:results}
\end{table}

Table~\ref{tab:comparison} compares our results to
the best spiking network results on these datasets in the literature.
The most significant recent results are from~\cite{Esser2016},
who implemented networks for a number of datasets on both one and eight
TrueNorth chips.
Their results are impressive,
but are difficult to compare with ours
since they use between 20 and 160 times more neurons.
We surpass a number of their one-chip results
while using an order of magnitude fewer neurons.
Furthermore, we demonstrate that our method scales to the large ILSVRC-2012 dataset,
which no other SNN implementation to date has done.
The most significant difference between our results
and that of \cite{Diehl2015} and \cite{Cao2014}
is that we use LIF neurons and can generalize to other neuron types,
whereas their methods (and those of \cite{Esser2016})
are specific to IF neurons.

\begin{table}
  \centering
  \begin{minipage}{\columnwidth}
    \begin{center}
    \begin{tabular}{|l|c|c|c|c|}\hline
      Dataset & \textbf{This Paper} & TN 1-chip & TN 8-chip & Best Other \\\hline\hline
      MNIST & 0.88\% (27k) & None & None & 0.88\% (22k)~\cite{Diehl2015} \\\hline
      SVHN & 6.08\% (27k) & 3.64\% (1M) & 2.83\% (8M) & None \\\hline
      CIFAR-10 & 16.46\% (50k) & 17.50\% (1M) & 12.50\% (8M) & 22.57\% (28k) \cite{Cao2014} \\\hline
      CIFAR-100 & 44.87\% (50k) & 47.27\% (1M) & 36.95\% (8M) & None \\\hline
      ILSVRC-2012 & 48.2\%, 23.8\% (493k)$^a$ & None & None & None\\\hline
    \end{tabular}
    \end{center}
    \vspace{-0.2em}
    \hspace{1em}$^a$ Results from the first 3072-image test batch.
  \end{minipage}
  \vspace{0.2em}
  \caption{
    Our error rates compared with
    recent results on the TrueNorth (TN) neuromorphic chip~\cite{Esser2016},
    as well as other best results in the literature.
    Approximate numbers of neurons are shown in parentheses.
    The TrueNorth networks use significantly more neurons than our networks
    (about 20$\times$ more for the 1-chip network
    and 160$\times$ more for the 8-chip network).
    The first number for ILSVRC-2012 (ImageNet) indicates the error for the top result,
    and the second number the more commonly reported top-5 result.
  }
  \label{tab:comparison}
\end{table}

We examined our methods in more detail on the CIFAR-10 dataset.
This dataset is composed of 60000 $32 \times 32$ pixel labelled images
from ten categories.
We used the first 50000 images for training and the last 10000 for testing,
and augmented the dataset by taking random $24 \times 24$ patches from the
training images
and then testing on the center patches from the testing images.
This methodology is similar to Krizhevsky et al. \cite{Krizhevsky2012},
except that they also used multiview testing where the classifier output
is the average output of the classifier run on nine random patches from
each testing image (increasing the accuracy by about 2\%).

Table~\ref{tab:mods} shows the effect of each modification
on the network classification error.
Rows 1-5 show that each successive modification
required to make the network amenable to running
in spiking neurons adds additional error.
Despite the fact that training with noise adds additional error
to the ANN,
rows 6-8 of the table show that in the spiking network,
training with noise pays off,
though training with too much noise is not advantageous.
Specifically, though training with $\sigma = 20$ versus $\sigma = 10$
decreased the error introduced when switching to spiking neurons,
it introduced more error to the ANN (Network 5),
resulting in worse SNN performance (Network 8).

\begin{table}
  \centering
  \begin{tabular}{|c|l|c|}\hline
    \# & Modification & CIFAR-10 error \\\hline\hline
    0 & Original ANN based on Krizhevsky et al. \cite{Krizhevsky2012} & 14.03\% \\\hline
    1 & Network 0 minus local contrast normalization & 14.38\% \\\hline
    2 & Network 1 minus max pooling & 16.70\% \\\hline\hline
    3 & Network 2 with soft LIF & 15.89\% \\\hline
    4 & Network 3 with training noise ($\sigma = 10$) & 16.28\% \\\hline
    5 & Network 3 with training noise ($\sigma = 20$) & 16.92\% \\\hline\hline
    6 & Network 3 ($\sigma = 0$) in spiking neurons & 17.06\% \\\hline
    7 & Network 4 ($\sigma = 10$) in spiking neurons & \textbf{16.46\%} \\\hline
    8 & Network 5 ($\sigma = 20$) in spiking neurons & 17.04\% \\\hline
  \end{tabular}
  \vspace{0.2em}
  \caption{Effects of successive modifications to CIFAR-10 error.
    We first show the original ANN based on \cite{Krizhevsky2012},
    and then the effects of each subsequent modification.
    Rows 6-8 show the results of running ANNs 3-5 in spiking neurons, respectively.
    Row 7 is the best spiking network, using a moderate amount of training noise.
  }
  \label{tab:mods}
\end{table}

\subsection{Efficiency}

Running on standard hardware,
spiking networks are considerably less efficient
than their ANN counterparts.
This is because ANNs are static,
requiring only one forward-pass through the network to compute the output,
whereas SNNs are dynamic,
requiring the input to be presented for a number of time steps
and thus a number of forward passes.
On hardware that can take full advantage of the sparsity that spikes provide---%
that is, neuromorphic hardware---%
SNNs can be more efficient than the equivalent ANNs, as we show here.

First, we need to compute the computational efficiency of the original network,
specifically the number of floating-point operations (flops) required to pass
one image through the network.
There are two main sources of computation in the image:
computing the neurons and computing the connections.
\begin{align}
  \text{flops} &= \frac{\text{flops}}{\text{neuron}} \times \text{neurons} +
                  \frac{\text{flops}}{\text{connection}} \times \text{connections}
\end{align}
Since a rectified linear unit is a simple max function,
it requires only one flop to compute (flops/neuron~=~1).
Each connection requires two flops, a multiply and an add (flops/connection~=~2).
We can determine the number of connections by ``unrolling'' each convolution,
so that the layer is in the same form as a locally connected layer.

To compute the SNN efficiency on a prospective neuromorphic chip,
we begin by identifying the energy cost
of a synaptic event ($E_{synop}$) and neuron update ($E_{update}$),
relative to standard hardware.
In consultation with neuromorphic experts,
and examining current reports of neuromorphic chips (e.g.~\cite{Merolla2014}),
we assume that
each neuron update takes as much energy as 0.25 flops ($E_{update} = 0.25$),
and each synaptic event takes as much energy as 0.08 flops ($E_{synop} = 0.08$).
(These numbers could potentially be much lower for analog chips, e.g.~\cite{Benjamin2014}.)
Then, the total energy used by an SNN to classify one image is
(in units of the energy required by one flop on standard hardware)
\begin{align}
  E_{SNN} = \left(E_{synop} \frac{\text{synops}}{\text{s}} +
            E_{update} \frac{\text{updates}}{\text{s}}\right) \times
            \frac{\text{s}}{\text{image}}
  \label{eqn:esnn}
\end{align}
For our CIFAR-10 network, we find that on average, the network has rates of
2,693,315,174 synops/s and 49,536,000 updates/s.
This results in $E_{CIFAR-10} = 45,569,843$,
when each image is presented for 200 ms.
Dividing by the number of flops per image on standard hardware,
we find that the relative efficiency of the CIFAR-10 network is 0.76,
that is it is somewhat less efficient.

Equation~\ref{eqn:esnn} shows that if we are able to lower
the amount of time needed to present each image to the network,
we can lower the energy required to classify the image.
Alternatively, we can lower the number of synaptic events per second
by lowering the firing rates of the neurons.
Lowering the number of neuron updates would have little effect on the
overall energy consumption since the synaptic events require the majority of the energy.

To lower the presentation time required for each input while maintaining accuracy,
we need to decrease the synapse time constant as well,
so that the information is able to propagate through the whole network
in the decreased presentation time.
Table~\ref{tab:efficiency} shows the effect of various alternatives
for the presentation time and synapse time constant
on the accuracy and efficiency of the networks for a number of the datasets.

\begin{table}
  \centering
  \begin{tabular}{|l|r|r|r|r|r|}\hline
    \multicolumn{1}{|c|}{Dataset} &
    \multicolumn{1}{|c|}{$\tau_s$ [ms]} &
    \multicolumn{1}{|c|}{$c_0$ [ms]} &
    \multicolumn{1}{|c|}{$c_1$ [ms]} &
    \multicolumn{1}{|c|}{Error} &
    \multicolumn{1}{|c|}{Efficiency} \\\hline\hline
    CIFAR-10 & 5 & 120 & 200 & 16.46\% & 0.76$\times$ \\\hline
    CIFAR-10 & 0 & 10 & 80 & 16.63\% & 1.64$\times$ \\\hline
    CIFAR-10 & 0 & 10 & 60 & 17.47\% & 2.04$\times$ \\\hline\hline
    MNIST & 5 & 120 & 200 & 0.88\% & 5.94$\times$ \\\hline
    MNIST & 2 & 40 & 100 & 0.92\% & 11.98$\times$ \\\hline
    MNIST & 2 & 50 & 60 & 1.14\% & 14.42$\times$ \\\hline
    MNIST & 0 & 20 & 60 & 3.67\% & 14.42$\times$ \\\hline\hline
    ILSVRC-2012 & 3 & 140 & 200 & 23.80\% & 1.39$\times$ \\\hline
    ILSVRC-2012 & 0 & 30 & 80 & 25.33\% & 2.88$\times$ \\\hline
    ILSVRC-2012 & 0 & 30 & 60 & 25.36\% & 3.51$\times$ \\\hline
  \end{tabular}
  \caption{Estimated efficiency of our networks on neuromorphic hardware,
    compared with traditional hardware.
    For all datasets, there is a tradeoff between accuracy and efficiency,
    but we find many configurations that are significantly more efficient
    while sacrificing little in terms of accuracy.
    $\tau_s$~is the synapse time constant,
    $c_0$~is the start time of the classification,
    $c_1$~is the end time of the classification
    (i.e. the total presentation time for each image).
  }
  \label{tab:efficiency}
\end{table}

Table~\ref{tab:efficiency} shows that for some datasets
(e.g. CIFAR-10 and ILSVRC-2012)
the synapses can be completely removed ($\tau_s = 0$ ms)
without sacrificing much accuracy.
Interestingly, this is not the case with the MNIST network,
which requires at least some measure of synapses to function accurately.
We suspect that this is because the MNIST network has much lower firing rates
than the other networks
(average of 9.67 Hz for MNIST, 148 Hz for CIFAR-10, 93.3 Hz for ILSVRC-2012).
This difference in average firing rates is also why the MNIST network is
significantly more efficient than the other networks.

\begin{figure}
  \centering
  \begin{tabular}{cc}
    CIFAR-10 ($\tau_s = 5$ ms) & CIFAR-10 ($\tau_s = 0$ ms) \\
    \includegraphics[width=0.4\columnwidth,clip=true,trim=2mm 0mm 10mm 8mm]{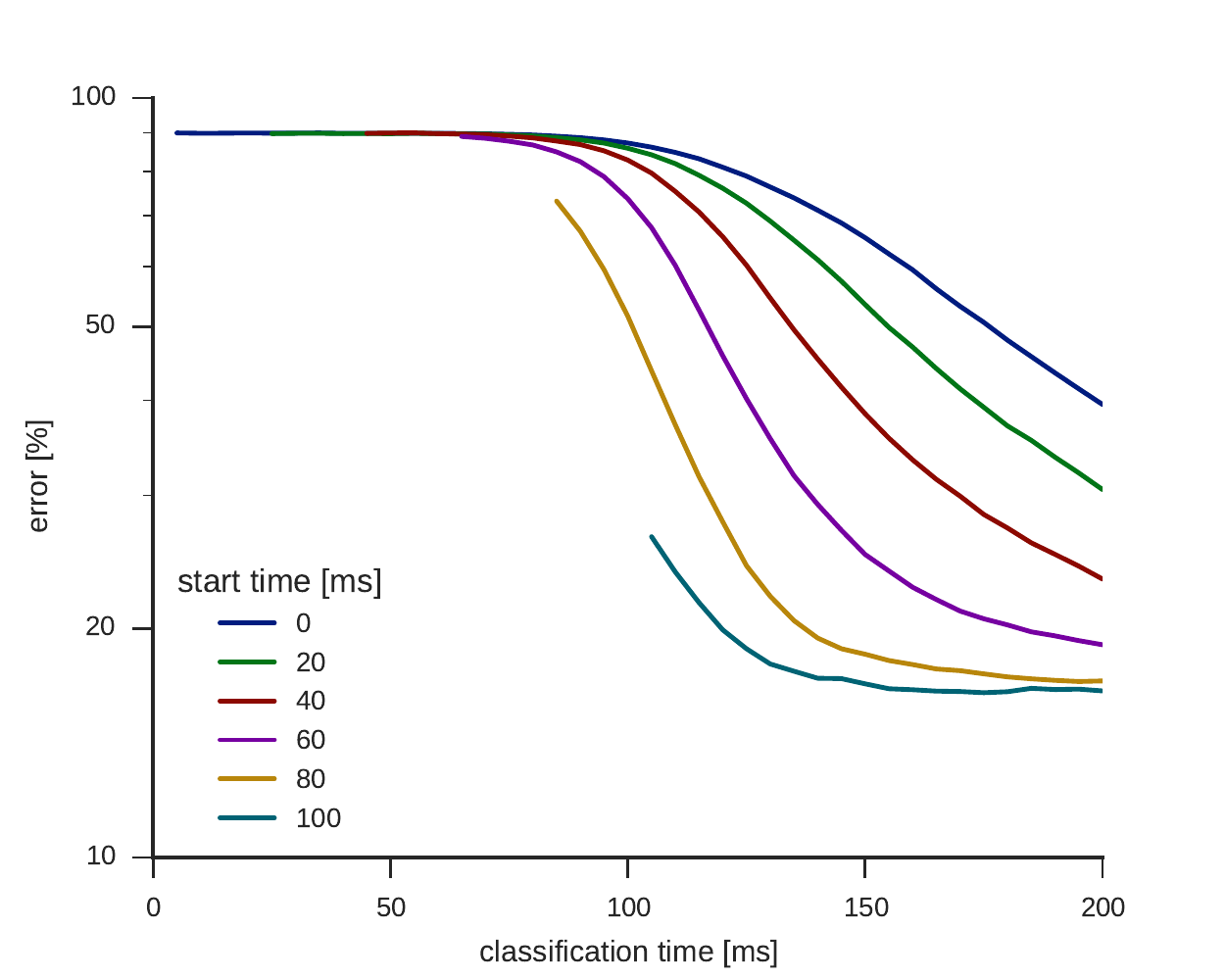} &
    \includegraphics[width=0.4\columnwidth,clip=true,trim=2mm 0mm 10mm 8mm]{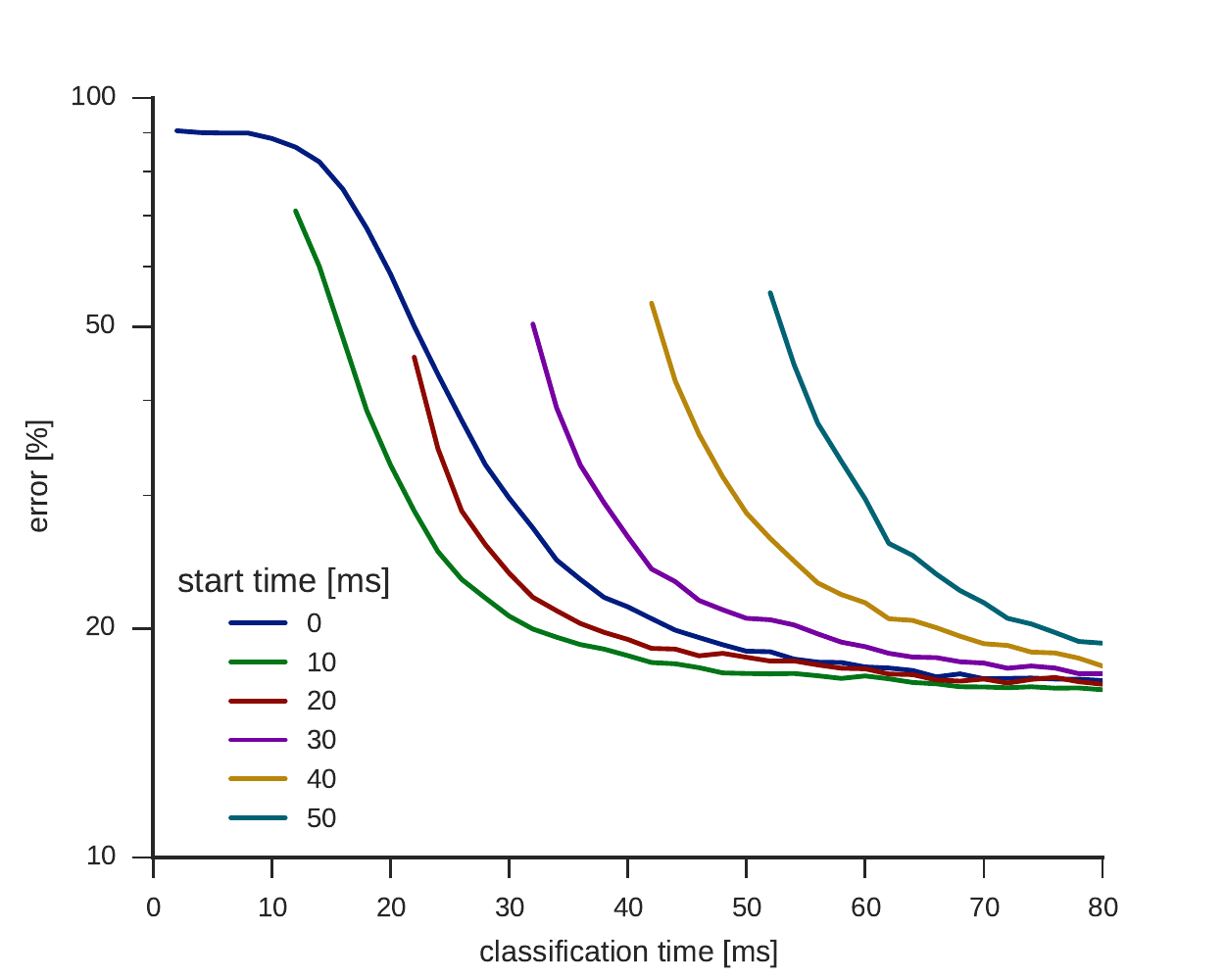} \\
    MNIST ($\tau_s = 2$ ms) & ILSVRC-2012 ($\tau_s = 0$ ms) \\
    \includegraphics[width=0.4\columnwidth,clip=true,trim=2mm 0mm 10mm 8mm]{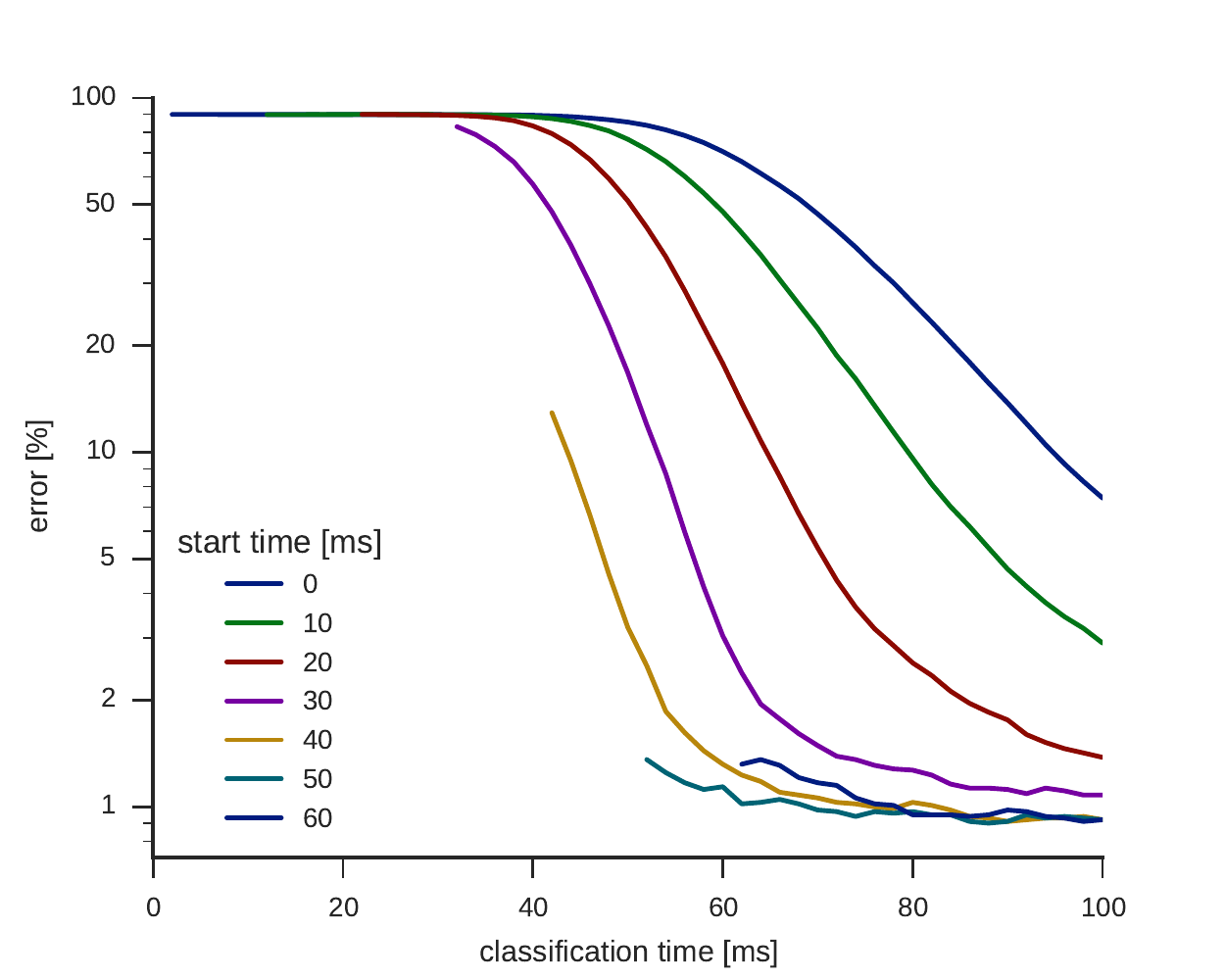} &
    \includegraphics[width=0.4\columnwidth,clip=true,trim=2mm 0mm 10mm 8mm]{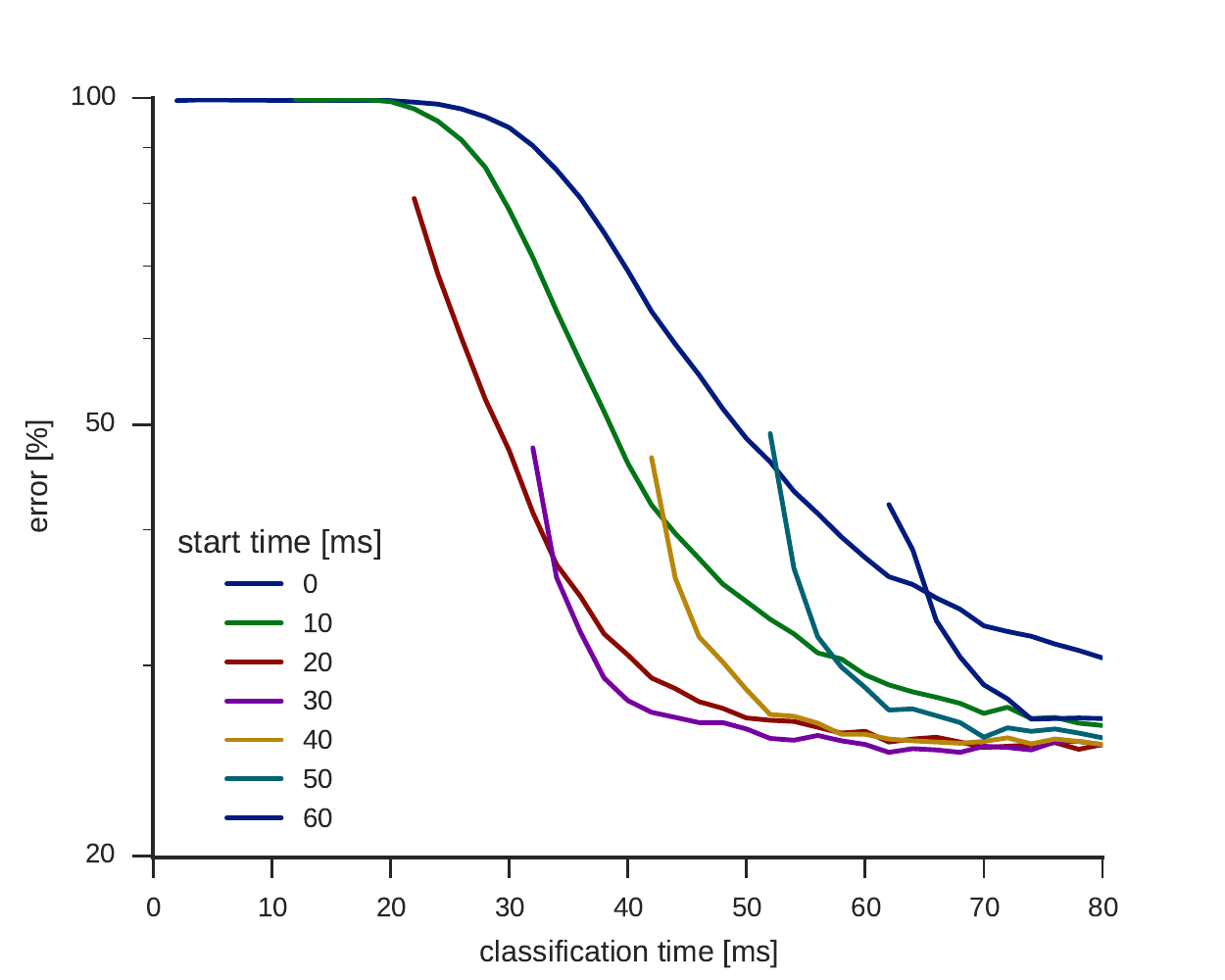} \\
  \end{tabular}
  \caption{Effects of classification time on accuracy.
    Individual traces show different starting classification times ($c_0$),
    and the x-axis the end classification time ($c_1$).
  }
  \label{fig:classtime}
\end{figure}

It is important to tune the classification time,
both in terms of the total length of time each example is shown for ($c_1$),
and when classification begins ($c_0$).
The optimal values for these parameters are very dependent on the network,
both in terms of the number of layers, firing rates, and synapse time constants.
Figure~\ref{fig:classtime} shows how the classification time affects accuracy
for various networks.

Given that the CIFAR-10 network performs almost as well with no synapses
as with synapses,
one may question whether noise is required during training at all.
We retrained the CIFAR-10 network with no noise and ran with no synapses,
but could not achieve accuracy better than 18.06\%.
This suggests that noise is still beneficial during training.

\section{Discussion}

Our results show that it is possible to train
accurate deep convolutional networks for image classification
without adding neurons,
while using more complex nonlinear neuron types---%
specifically the LIF neuron---%
as opposed to the traditional rectified-linear or sigmoid neurons.
We have shown that networks can be run in spiking neurons,
and training with noise decreases the amount of error introduced
when running in spiking versus rate neurons.
These networks can be significantly more energy-efficient than traditional ANNs
when run on specialized neuromorphic hardware.

The first main contribution of this paper is to demonstrate that
state-of-the-art spiking deep networks can be trained with LIF neurons,
while maintaining high levels of classification accuracy.
For example, we have described the first large-scale SNN
able to provide good results on ImageNet.
Notably, all other state-of-the-art methods use integrate-and-fire (IF) neurons
\cite{Cao2014, Diehl2015, Esser2016},
which are straightforward to fit to the rectified linear units commonly used
in deep convolutional networks.
We show that there is minimal drop in accuracy when converting from ANN to SNN.
We also examine how classification time affects accuracy and energy-efficiency,
and find that networks can be made quite efficient
with minimal loss in accuracy.

By smoothing the LIF response function so that its derivative remains bounded,
we are able to use this more complex and nonlinear neuron
with a standard convolutional network trained by backpropagation.
Our smoothing method is extensible to other neuron types,
allowing for networks to be trained for neuromorphic hardware with
idiosyncratic neuron types (e.g.~\cite{Benjamin2014}).
We found that there was very little error introduced
by switching from the soft response function to the hard response function
with LIF neurons for the amount of smoothing that we used.
However, for neurons with harsh discontinuities that require more smoothing,
it may be necessary to slowly relax the smoothing
over the course of the training
so that, by the end of the training, the smooth response function
is arbitrarily close to the hard response function.

The second main contribution of this paper is to demonstrate
that training with noise on neuron outputs
can decrease the error introduced when transitioning to spiking neurons.
The error decreased by 0.6\% overall on the CIFAR-10 network,
despite the fact that the ANN trained without noise performs better.
This is because noise on the output of the neuron simulates the variability
that a spiking network encounters when filtering a spike train.
There is a tradeoff between training with too little noise,
which makes the SNN less accurate,
and too much noise, which makes the initially trained ANN less accurate.

These methods provide new avenues for translating traditional ANNs
to spike-based neuromorphic hardware.
We have provided some evidence that such implementations can be
significantly more energy-efficient than their ANN counterparts.
While our analyses only consider static image classification,
we expect that the real efficiency of SNNs will become apparent
when dealing with dynamic inputs (e.g. video).
This is because SNNs are inherently dynamic,
and take a number of simulation steps to process each image.
This makes them best suited to processing dynamic sequences,
where adjacent frames in the video sequence are similar to one another,
and the network does not have to take time to constantly ``reset''
after sudden changes in the input.

Future work includes experimenting with lowering firing rates
for greater energy-efficiency.
This could be done by changing the neuron refractory period $\tau_{ref}$
to limit the firing below a particular rate,
optimizing for both accuracy and low rates,
using adapting neurons,
or adding lateral inhibition in the convolutional layers.
Other future work includes implementing max-pooling
and local contrast normalization layers in spiking networks.
Networks could also be trained offline as described here
and then fine-tuned online using an STDP rule \cite{Nessler2013, Bekolay2013}
to help further reduce errors associated with converting from
rate-based to spike-based networks,
while avoiding difficulties with training
a network in spiking neurons from scratch.


{\small

}

\end{document}